\documentclass[10pt,twocolumn]{ICCAS2022}
 
\usepackage{diagbox}
\usepackage{enumitem}
\usepackage{booktabs}
\usepackage{multirow}
\usepackage{etoolbox}
\usepackage{pgf}
\usepackage{colortbl}
\usepackage{amsmath}
\usepackage{tikz}
\usepackage[ruled,linesnumbered]{algorithm2e}
\usepackage{url}
\usepackage[hidelinks]{hyperref}
\hypersetup{
    colorlinks,                 %Colors links instead of ugly boxes
    linkcolor=black,            %Color of internal links
    citecolor={blue!50!black},  %Color of citations
    urlcolor={blue!80!black}    %Color for external hyperlinks
}

\begin{document}

\title{Robotic Dough Shaping}

% \author{First A. Author${}^{1}$ and Second B. Author${}^{2*}$ }
\author{Jan Ondras${}^{*}$, Di Ni, Xi Deng, Zeqi Gu, Henry Zheng and Tapomayukh Bhattacharjee}

\affils{ Cornell University, NY, USA (\{jo339, dn273, xd93, zg45, hz265, tb557\}@cornell.edu) {\small${}^{*}$ Corresponding author}}
% \affils{ ${}^{1}$Department of Electrical Engineering, Hankook University, \\
% Seoul, 13391, Korea (first@hankook.ac.kr) \\
% ${}^{2}$Department of Mechanical Engineering, Hankook University, \\
% Seoul, 13391, Korea (second@hankook.ac.kr) {\small${}^{*}$ Corresponding author}}

%\thanks{ \noindent
%   This paper is supported by my funding agencies.
%  }

\abstract{
    % It is recommended that your abstract contain 150-200 words.
Robotic manipulation of deformable objects gains great attention due to its wide applications including medical surgery, home assistance, and automatic food preparation. 
The ability to deform soft objects remains a great challenge for robots due to difficulties in defining the problem mathematically. 
% Di: I think it;s important to address the motivation and challenges here. I'm just talking nonsense here. Someone knows better about the background please polish it... 
In this paper, we address the problem of shaping a piece of dough-like deformable material into a 2D target shape presented upfront. 
We use a 6 degree-of-freedom WidowX-250 Robot Arm equipped with a rolling pin and information collected from an RGB-D camera and a tactile sensor. 
We present and compare several control policies, including a dough shrinking action, %besides the expansion action
%that generalize to 
in extensive experiments across three kinds of deformable materials and across three target dough shape sizes, achieving the intersection over union (IoU) of 0.90. 
% Our Roll Dough GUI Application
Our results show that: i) rolling dough from the highest dough point is more efficient than from the 2D/3D dough centroid; ii) it might be better to stop the roll movement at the current dough boundary as opposed to the target shape outline; iii) the shrink action might be beneficial only if properly tuned with respect to the expand action; and iv) the \textit{Play-Doh} material is easier to shape to a target shape as compared to \textit{Plasticine} or \textit{Kinetic sand}.
Video demonstrations of our work are available at 
\href{https://youtu.be/ZzLMxuITdt4}{https://youtu.be/ZzLMxuITdt4}
% \url{https://youtu.be/ZzLMxuITdt4}
% https://youtu.be/ZzLMxuITdt4
% what problem
% methods overview
% results a sample iou
% findings/insights
}


\maketitle

%-----------------------------------------------------------------------
\section{INTRODUCTION} \label{sec:introduction}
%%%%%%%%%%%%%%%%%%%%%%%%%%%%%%%%%%%%%%%%%%%%%%%%%%%%%%%%%%%%%%%%%%%%%%%%%%%%%%%%
% What to write here:
%   What is the problem? 
%   Why is it interesting and important? 
%   Why is it hard? 
%   Why hasn't it been solved before?
%%%%%%%%%%%%%%%%%%%%%%%%%%%%%%%%%%%%%%%%%%%%%%%%%%%%%%%%%%%%%%%%%%%%%%%%%%%%%%%%
% Can inspire from papers by Shuran Song, Oliver Kroemer
Robotic manipulation of deformable objects has applications in various domains, including robotic surgery, home
robotic solutions, and automated food preparation. A ubiquitous challenge throughout this area of research is that robotic interactions with deformable materials are
highly complex, often nonlinear, and difficult to model, which further increases the difficulty of action selection and real-time computation. Another challenge is the gap between simulation and real world environments, especially amplified for deformable objects that are often impossible to model perfectly.

% What we did & Contributions of the paper - what is new?
The particular task that this work focuses on is shaping elastoplastic solid 3D deformable objects to a predefined target 2D shape on the horizon axis. To make the problem tractable while keeping the solution generalizable, we design a simple action space to roll out a piece of dough against a horizontal flat surface and track the dough's shape progression with a low-dimensional feature space. We apply and test our methods directly in real world, skipping the simulation phase to avoid the sim-to-real gap. For the hardware execution, we design an end-effector attachment to the robot arm. Our main contributions are:
\begin{itemize}[leftmargin=10pt]
    \item  Efficient heuristic dough shaping methods with multimodal (vision and 3D point clouds) information processing, and their evaluations on three kinds of materials with different energy density. 
    % We further investigated a physically-based differentiable rendering method to see whether it helps to improve the performance.
    \item Roll Dough GUI Application\footnote{Available at \href{https://github.com/jancio/Robotic-Dough-Shaping}{https://github.com/jancio/Robotic-Dough-Shaping}} supporting various configurations and methods.
    % Available at 
    % \small{\url{https://github.com/jancio/Robotic-Dough-Shaping}}\normalsize
    % \hyperlink{https://github.com/jancio/Robotic-Dough-Shaping}{https://github.com/jancio/Robotic-Dough-Shaping}
    % \href{https://github.com/jancio/Robotic-Dough-Shaping}{https://github.com/jancio/Robotic-Dough-Shaping}
    \item Material properties analysis with a tactile sensor which can inform the robot control strategies in further work.
\end{itemize}
% Structure

% The rest of this paper is organized as follows. 
%%%% Section~\ref{sec:problem_definition} defines the problem addressed by our work,
% Section \ref{sec:related_work} reviews the related work in the field and 
% Section \ref{sec:methods} describes the methodology.
% Performed experiments and obtained results are presented and discussed in Section \ref{sec:experiments}. 
% Section \ref{sec:conclusion} summarizes our work and suggests directions for further research.

% \input{sections/2_problem_definition}
\section{RELATED WORK} \label{sec:related_work}
% from notes

% There is a plethora of previous work on deformable object manipulation, ranging from manipulation of cloths~\cite{chi2022iterative, ha2022flingbot} to food~\cite{sawhney2020playing}. 
% Most previous work on dough-like deformable object manipulation focused on simulated environments~\cite{huang2021plasticinelab, lin2021diffskill}. 
% % Only, the work~\cite{}  deformable manipulation using a soft endeffector 
% The work~\cite{matl2021deformable} is most closely related to ours. 
% They developed a novel soft end-effector to roll elasto-plastic dough into different lengths using model-based reinforcement learning. 
% They also demonstrated the method in real robot experiments.
% demonstrate plasticine manipulation with a real robot

\subsection{Deformable Object Manipulation}
Compared to rigid bodies, the new challenges introduced by Deformable Object Manipulation (DOM) can be generally grouped by three aspects: 
i) designing new devices and algorithms for more accurate sensing and perception~\cite{hui2017visual,guler2015estimating,nair2017combining,yan2020learning,rouhafzay2020transfer}, ii) modeling the complex deformation~\cite{li2015regrasping,zhu2021vision,navarro2017fourier,lagneau2020automatic},
and iii) the high degree of freedom in planning~\cite{mcconachie2020manipulating,ramirez2014motion,alonso2015local} and control~\cite{mitsioni2019data,su2019improved} of manipulation actions. 
When conducting experiments, the objects usually have a relatively simple shape: either linear such as ropes~\cite{yan2020learning,sundaresan2020learning} and cables~\cite{zhu2018dual}, or planar objects such as cloth~\cite{chi2022iterative,ha2022flingbot} and gauze~\cite{thananjeyan2017multilateral}. In this work, we focus on exploring i) and iii) for dough-like object, which is relatively less explored yet very common in real life.

\subsection{Policy Learning}
Many existing works use reinforcement learning (RL) to approach the task.  PlasticineLab~\cite{huang2021plasticinelab} establishes one of the first deformable object manipulation benchmarks and evaluates popular RL algorithms on them. The work~\cite{matl2021deformable} is perhaps the most similar to ours as it also deals with dough shaping with an end-effector, and it requires a random exploration period to learn the policy before solving the task. 
% Due to time constraints, 
We research on heuristic policies, which turn out to be still very efficient. \cite{ha2022flingbot} proposes an RL algorithm that shares the same intuition as ours: dynamic manipulation with delta physics. Instead of deciding the next action based on the holistic view of current state, another perspective is to base it on how far the current state is from the target state, thus the "delta". This is particularly useful for deformable objects where the generalization and state characterization are much harder than for regular rigid bodies.

\subsection{Tactile Sensing}
% Two subsections:
% tactile sensing with robots
% tactile sensing of deformable materials
Besides the visual input, a robot is desired to have human-like tactile sensing abilities for contact-rich tasks. The ability to sense through physical contact is especially important when the vision information is not available (limited light source) or under large distortion (large reflection or underwater) . The study of tactile sensing has started more than 30 years ago~\cite{nicholls1989survey}. The tactile information is generally used in two ways: i) haptic data can be used independently for object identification~\cite{xu2013tactile,allen1988integrating,allen1988haptic,jamali2011majority}, or ii) the contact information can be combined with other inputs, such as the visual inputs, so that robots can learn from multi-modal representations~\cite{fazeli2019see,romano2011human,lee2020making}. 
Using multimodal fusion, tactile sensing was first integrated with other sensor inputs based on handcrafted features~\cite{romano2011human}. 
For example, the work~\cite{xu2013tactile} demonstrates the sensing of compliance, texture, and thermal properties, based on which Bayesian exploration was used for object identification. 
More often, tactile information was used with vision input. 
For instance, Lee et al.~\cite{lee2020making} used both vision and tactile information to train a self-supervised model for a peg insertion task.

% Tell what is new
% However, to the best of our knowledge, there have not been any real robot 

\vspace{-5mm}
\section{METHODS} \label{sec:methods}
In this section, we present 
the hardware setup (Section~\ref{sec:methods:hardware} ~~), 
initial and target dough shape definition (Section~\ref{sec:methods:init_target_shape} ~~ ),
Roll Dough Algorithm including various control methods (Section \ref{sec:methods:roll_dough_algo} ~),
and approach to the tactile sensing of the dough (Section \ref{sec:methods:tactile_sensing} ~ ).

%%%%%%%%%%%%%%%%%%%%%%%%%%%%%%%%%%%%%%%%%%%%%%%%%%%%%%%%%%%%%%%%%%%%%%%%%%%%%%%%
\subsection{Hardware setup}\label{sec:methods:hardware}

\begin{figure}[ht]
    \centering
    \hspace*{-0.6ex}%
    \begin{tikzpicture}[x=0.45\textwidth,y=0.1\textwidth,every text node part/.style={align=center}]
        \node at (-0.1, 0.0){};
        \node at (0.0, 0.0){
        \includegraphics[height=0.28\textwidth]{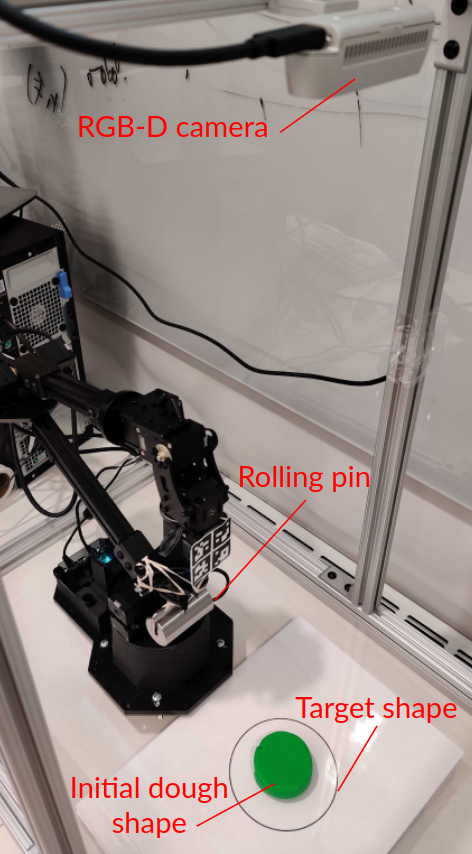}};
        \node at (0.53, 0.0){
         \includegraphics[trim={0 0 1.9cm 0},clip,height=0.28\textwidth]{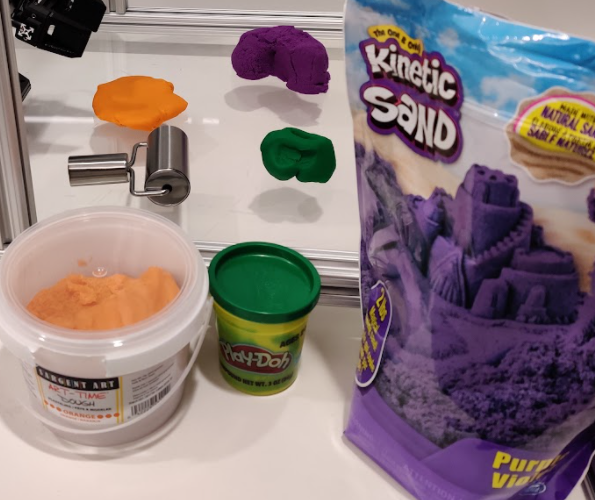}};
    \end{tikzpicture}
    \vspace{-5mm}
    \caption{
    \textbf{Left:} Robot and camera setup: 
    6 degree-of-freedom WidowX-250 Robot Arm equipped with a steel rolling pin and Intel RealSense D435i RGB-D camera mounted 60 cm above the robot workspace. 
    \textbf{Right:} Dough-like deformable materials considered: \textit{Play-Doh} (green), \textit{Plasticine} (orange), \textit{Kinetic sand} (purple).}
    \label{fig:hw_setup_materials}
    \vspace{-2mm}
\end{figure}

% \subsubection{Robot and camera}
As a robotic platform, we used a 6 degree-of-freedom WidowX-250 Robot Arm that we equipped with a rolling pin. 
For perception, we utilized the Intel RealSense D435i RGB-D camera mounted 60 cm above the robot workspace. 
The whole setup is shown in Figure~\ref{fig:hw_setup_materials} (left).
\vspace{-1mm}

% \subsubection{Dough-like materials}
We considered the following three dough-like deformable materials with different properties, all shown in Figure~\ref{fig:hw_setup_materials}~(right).
\begin{itemize}[leftmargin=10pt]
    \item \textbf{Play-Doh} -- easiest to deform, dries out fastest.
    \item \textbf{Plasticine} -- harder to deform, dries out slowly.
    \item \textbf{Kinetic sand} -- easy to deform only if the deformation is slow, shares some properties of granular media, does not dry out.
\end{itemize}
% \begin{figure}[ht]
%     \centering
%     \includegraphics[width=0.485\textwidth]{fig/methods/materials.png}
%     \caption{}
%     \label{fig:materials}
% \end{figure}

%%%%%%%%%%%%%%%%%%%%%%%%%%%%%%%%%%%%%%%%%%%%%%%%%%%%%%%%%%%%%%%%%%%%%%%%%%%%%%%%
\subsection{Initial and Target Dough Shape}\label{sec:methods:init_target_shape}

We standardized the initial dough shape using a plastic cylindrical mold with the diameter of 5.6 cm and height of 1.6 cm, placed at a fixed location with respect to the target shape (Figure~\ref{fig:hw_setup_materials} (left)). 
This configuration ensured same initial conditions for all our experiments. 
However, our methods can handle a general case when this is not satisfied. 

We considered three circular 2D target shapes with diameters 3.5, 4.0, and 4.5 inch, denoted by $T_{3.5}$, $T_{4.0}$, and $T_{4.5}$ respectively. 
% , placed at any location within the workspace
Each target shape was printed on a paper and placed just below the transparent workspace plate (Figure~\ref{fig:hw_setup_materials}~(left)).

% possibly add a figure with initial and target shape (slide 9)

%%%%%%%%%%%%%%%%%%%%%%%%%%%%%%%%%%%%%%%%%%%%%%%%%%%%%%%%%%%%%%%%%%%%%%%%%%%%%%%%
\subsection{Roll Dough Algorithm}\label{sec:methods:roll_dough_algo}

To roll the dough, we developed an iterative procedure presented in Algorithm~\ref{algo:roll_dough} and implemented as a GUI application (see Appendix \ref{sec:appendix:app} ~ for screenshots) supporting various configurations and methods available at 
\href{https://github.com/jancio/Robotic-Dough-Shaping}{https://github.com/jancio/Robotic-Dough-Shaping}
% First, we provide its high-level description and then provide details on roll start point calculation method, roll end point calculation method, and an extension with shrink action.

At the beginning, we detect a circular 2D target shape in an RGB image using the OpenCV library~\cite{opencvlibrary} (line 2). 
In an iterative manner, we then similarly detect contours of the current dough shape (lines 3 and 8),
evaluate the intersection over union (IoU) metric between the current dough projected 2D shape and the target 2D shape (lines 4 and 9), 
and calculate the next roll start point $S$ and end point $E$ (lines 5 and 10). Given these 3D points, the robot arm with the rolling pin touches the dough at point $S$ and performs the roll action to point $E$ (line 7).
% The algorithm terminates when either the time $t$ reaches the maximum time limit $T_{max}$ or a minimum IoU $IoU_{min}$ is reached (line 6).
The algorithm terminates when either the maximum time limit $T_{max}$ or the minimum IoU $IoU_{min}$ is reached (line 6).
\vspace{-2mm}
\begin{algorithm}[ht]
    % \SetAlgoVlined
    \SetAlgoNoLine
    \SetAlgoNoEnd
    \caption{Roll Dough Algorithm}
    \label{algo:roll_dough}
\hspace{1mm}        $t \gets 0$\;
\hspace{1mm}        $targetShp$ $\gets$ captureTargetShape()\;
\hspace{1mm}        $currentShp$ $\gets$ captureCurrentShape()\;
\hspace{1mm}        $IoU$ $\gets$ evaluate($currentShp$, $targetShp$)\;
\hspace{1mm}        $S$, $E$ $\gets$ plan($currentShp$, $targetShp$)\;
\hspace{1mm}        \While{$t < T_{max}$ \textbf{and} $IoU < IoU_{min}$}{
\hspace{5mm}        	roll($S$, $E$)\;
\hspace{5mm}            $currentShp$ $\gets$ captureCurrentShape()\;
\hspace{5mm}            $IoU$ $\gets$ evaluate($currentShp$, $targetShp$)\;
\hspace{5mm}            $S$, $E$ $\gets$ plan($currentShp$, $targetShp$)\;
        }
\end{algorithm}
\vspace{-3mm}

% Alternative
% \begin{algorithm}[ht]
% \caption{Roll Dough Algorithm}
% \label{algo:roll_dough}
% \begin{algorithmic}[1]
%     \State $t \gets 0$
%     \State $targetShape$ $\gets$ captureTargetShape()
%     \State $currentShape$ $\gets$ captureCurrentShape()
%     \State $IoU$ $\gets$ evaluate($currentShape$, $targetShape$)
%     \While{$t < T_{max}$ \textbf{and} $IoU < IoU_{min}$}
%     	\State roll($S$, $E$)
%         \State $currentShape$ $\gets$ captureCurrentShape()
%         \State $IoU$ $\gets$ evaluate($currentShape$, $targetShape$)
%         \State $S$, $E$ $\gets$ plan($currentShape$, $targetShape$)
%     \EndWhile
% \end{algorithmic}
% \end{algorithm}

In the following subsections, we provide details on the roll start point calculation method (Section \ref{sec:methods:roll_dough_algo:start_point} ~ ), roll end point calculation method (Section \ref{sec:methods:roll_dough_algo:end_point} ~ ), and an extension with a shrink action (Section \ref{sec:methods:roll_dough_algo:shrink_action} ~ ).

\subsubsection{Roll Start Point Method}\label{sec:methods:roll_dough_algo:start_point}
We choose the roll start point $S$ as a point from where the dough should to be distributed to other areas where it is currently missing.
As such, we consider the following four roll start point methods.
\begin{itemize}[leftmargin=10pt]
    \item \textbf{Centroid-2D} -- 
    % Given a contour of the current dough shape, 
    % We calculate the start point 
    We calculate the x and y coordinates $(S_x, S_y)$ of the start point $S$ as a 2D geometric center of the current dough shape.
    Using K-Nearest Neighbors we then find three closest points (by their x and y coordinates) in the 3D point cloud of the current dough shape and set the z coordinate $S_z$ to be the average z coordinate of these points.
    
    \item \textbf{Centroid-3D} -- 
    % Utilizing the depth information from the RGBD camera.
    Utilizing the whole dough point cloud data, we calculate the start point $S$ as a 3D location of the center of mass of the dough (assuming constant material density).

    \item \textbf{Highest-Point} -- 
    We set the start point $S$ to be the highest point in the dough point cloud.
    
    \item \textbf{Differential-Inverses-Renderings} -- 
    One intuitive approach is to differentiate a loss based on distribution difference between current point cloud and target shape.
    Before the iteration starts, we discretize the target shape as 3D point clouds, then back propagate the density aware chamfer loss \cite{wu2021densityaware} to each vertex in the current point cloud,
    \vspace{-2mm}
    \begin{align}
        L_{DCD} &= \frac{1}{2}(\frac{1}{|S_1|}\sum_{x\in S_1} (1 - \frac{1}{n_{\hat{y}}} e^{-\alpha || x - \hat{y}||}) \\ &+ \frac{1}{|S_2|}\sum_{y\in S_2} (1 - \frac{1}{n_{\hat{x}}} e^{-\alpha || y - \hat{x}||}))
    \end{align}
    with $\alpha = 1000$, between the target point distribution $S_1$ and current point cloud distribution $S_2$ to get point wise gradient. There are several ways to plan the movement direction from this gradient: using the mean value of the gradient or using the gradient with the maximum magnitude. We chose to use the gradient with largest magnitude because the mean gradient could have 0 magnitude in the xy plane if the pointcloud is perfectly aligned in the center. 
    % \begin{enumerate}
    %     \item Inverse in 2D.  The contour of the shape is described by a closed path $\path = v_0v_1v_2\dots v_0$. We define a renderer $I(\path)$ that take the path as input and output an silhouette images. The target image $\target$ can be easily defined by user. Then the differentiatiable renderer optimize $I(\path)$ towards $\target$ by minimising a loss function. The step is recorded an used as input of planning of a sequence of robot movement. During the manipulation, the RGB camera can get captures of the dough (from top view) and use it as a feedback to correct the planning. This method requires using the height information to determine where to start the pushing.
    %     \item Inverse in 3D. The object is initialised as triangle mesh $\mesh$, we define a physically-based renderer [psdr-cuda] $I(\mesh)$ that take the mesh, a point light colocated with camera, and a lambertian surface reflectance model as input and output RGB image of the object viewed from camera. The process is similar as the 2D version, but this one has more point defined and don't need additional height information on where to put the rolling pin.
    % \end{enumerate}
    % This method requires calibration of camera and light.
\end{itemize}

% \begin{figure}
%     \centering
%     \includegraphics[width=0.51\textwidth,page=1]{fig/dough_fig.pdf}
%     \caption{An example of RGB-Center method (top left). Two Usage of rolling pin for expanding and shrinking (bottom left). Illustration of differentiable rendering on 2D (right)}
%     \label{fig:inverse}
% \end{figure}

\subsubsection{Roll End Point Method}\label{sec:methods:roll_dough_algo:end_point}
\begin{figure}[ht]
    \centering
    \hspace*{-0.8ex}%
    \begin{tikzpicture}[x=0.46\textwidth,y=0.1\textwidth,every text node part/.style={align=center}]
        \node at (-0.1, 0.0){ };
        \node at (0.0, -0.85){(a)};
        \node at (0.33, -0.85){(b)};
        \node at (0.66, -0.85){(c)};
        \node at (0.0, 0.0){
        \includegraphics[trim={2.75cm 0 0 0},clip,height=0.14\textwidth]{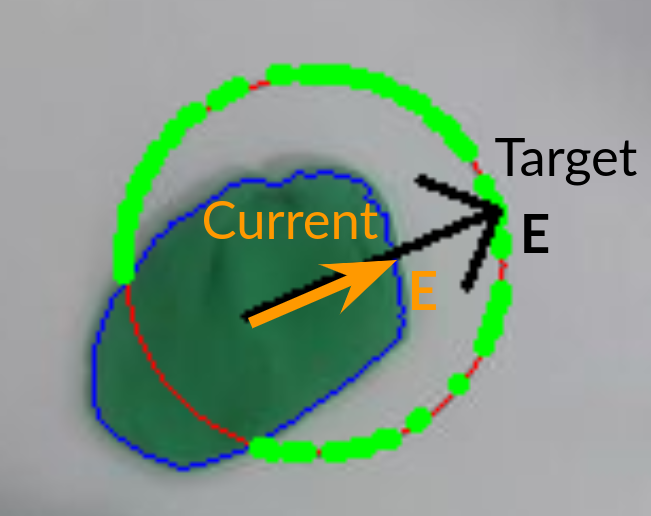}};
        \node at (0.517, 0.0){
        \includegraphics[height=0.14\textwidth]{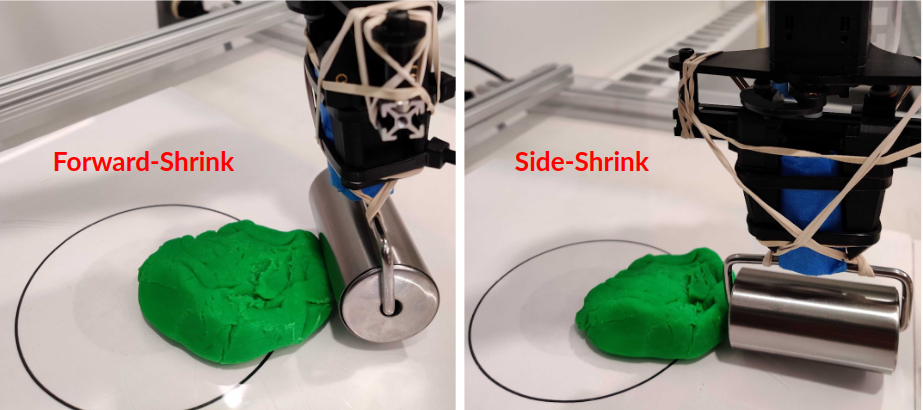}};
    \end{tikzpicture}
    \vspace{-6mm}
    \caption{Roll end point methods (a): 
        \textbf{Target} point $E$ is located on the target shape outline (red) 
        % with the largest gap between 
        and the green points denote all candidate points.
        \textbf{Current} point $E$ is located on the current dough shape outline (blue). 
        The directions where the dough is outside of the target shape are ignored. 
        Shrink action variants (b-c): 
        \textbf{Forward-Shrink}: the dough is pushed by a rolling pin in its usual roll orientation.
        \textbf{Side-Shrink}: the dough is pushed by the side of the rolling pin. This can also simulate a use of another tool, such as spatula, to perform the shrink action.
    }
    \label{fig:end_method_and_shrink_method}
    % \label{fig:shrink_method}
    \vspace{-6mm}
\end{figure}
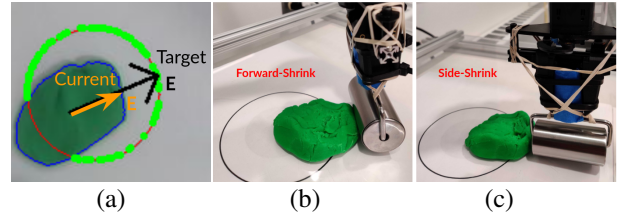

We roll the dough from the start point $S$ in the direction $\vec{d}$ in which there is a largest gap between the current and target shape, i.e., we employ the \textit{"minimize the largest error first"} heuristic. We ignore candidate directions where the dough is outside of the target shape. 
To set the roll end point $E$ in this direction, we consider the following two variants, also illustrated in Figure~\ref{fig:end_method_and_shrink_method} (a).
\begin{itemize}[leftmargin=10pt]
    \item \textbf{Target} -- We set the x and y coordinates ($E_x$, $E_y$) of the roll end point $E$ to be a 2D intersection point of the vector $\vec{d}$ and the target shape outline.
    \item \textbf{Current} -- We set ($E_x$, $E_y$) to be a 2D intersection point of the vector $\vec{d}$ and the current dough %shape 
    outline. %point $E$ is located on the current shape outline
\end{itemize}
In both methods, the z coordinate $E_z = S_z$.

\subsubsection{Shrink Action}\label{sec:methods:roll_dough_algo:shrink_action}
% In addition to expanding the dough, we consider a variant of the algorithm which allows a shrink action to . 
In theory, the roll end point methods should not spread the dough outside of the target shape. However, the dough can be moved there accidentally due to any sort of inaccuracies and the above-proposed Roll Dough Algorithm does not explicitly rectify the dough shape in such cases to increase IoU. 
% While rolling a piece of dough it can accidentally spread outside of the target shape. 
We thus also consider a variant of the Roll Dough Algorithm which allows a shrink action to correct the dough shape and achieve higher IoU in such cases.
% The shrink action is performed only 
At each iteration of the Roll Dough Algorithm, if any point on the current dough shape contour is found to be outside of the target shape, we execute the shrink action instead of the roll action. 
We set the start point of the shrink action to be the furthest point on the current dough shape contour outside of the target shape in the direction towards the target shape center,
and the end point to be the point on the target shape outline in this direction. The z coordinate of both start and end points is set to touch the workspace plate.

We consider the following two variants of the shrink action, also illustrated in Figure~\ref{fig:end_method_and_shrink_method} (b-c).
\begin{itemize}[leftmargin=10pt]
    \item \textbf{Forward-Shrink} -- The dough is pushed by a rolling pin in its usual roll orientation.
    \item \textbf{Side-Shrink} -- The dough is pushed by the side of the rolling pin. This can also simulate a use of another tool, such as spatula, to perform the shrink action.
\end{itemize}
We further denote the original variant without the shrink action as \textit{Shrink-Disabled}.

% \begin{figure}[ht]
%     \centering
%     \includegraphics[width=0.485\textwidth]{fig/methods/shrink.png}
%     \caption{Shrink action variants: 
%         \textbf{Forward-Shrink} -- The dough is pushed by a rolling pin in its usual roll orientation.
%         \textbf{Side-Shrink} -- The dough is pushed by the side of the rolling pin. This can also simulate a use of another tool, such as spatula, to perform the shrink action.
%     }
%     \label{fig:shrink_method}
% \end{figure}

    % This variant extends each of the previous variants by allowing the shrinking action in addition to the expansion action. 
    % The shrinking action is performed using the side edge of the rolling pin.
% he robotic arm distributes dough towards the target shape either by expanding or, if allowed, shrinking the dough. 

%%%%%%%%%%%%%%%%%%%%%%%%%%%%%%%%%%%%%%%%%%%%%%%%%%%%%%%%%%%%%%%%%%%%%%%%%%%%%%%%
\subsection{Dough Tactile Sensing}\label{sec:methods:tactile_sensing}
Tactile sensors are capable of detecting a rich range of information including stiffness, compliance and thermal properties. In this paper, we implemented a resistance based strain gauge, which is programmed to give feedback on the stiffness of the three types of dough. This information can be used to determine the planning of force amplitude, rolling speed and rolling iteration numbers. 

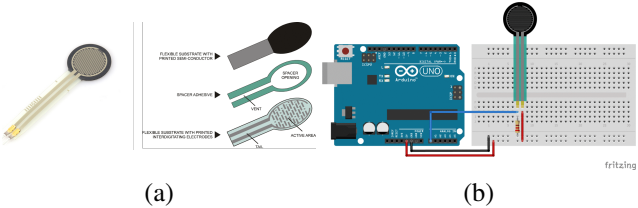
\begin{figure}[ht]
    \centering
    \hspace*{-3ex}%
    \begin{tikzpicture}[x=0.5\textwidth,y=0.1\textwidth,every text node part/.style={align=center}]
    \node at (-0.1, 0.0){ };
    \node at (0.0, -0.85){(a)};
    \node at (0.5, -0.85){(b)};
    \node at (0.0, 0.0){\includegraphics[trim={1.2cm 1.5cm 0 0},clip,width=0.245\textwidth]{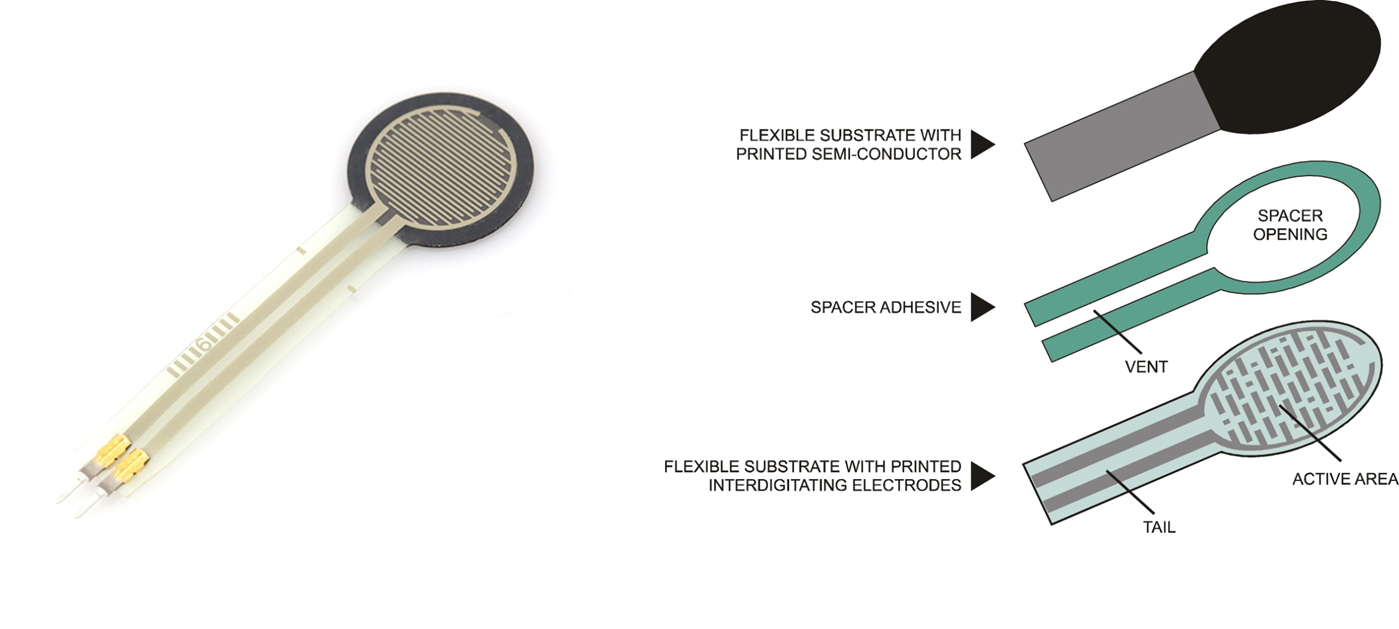}};
    \node at (0.5, 0.0){
    \includegraphics[width=0.245\textwidth]{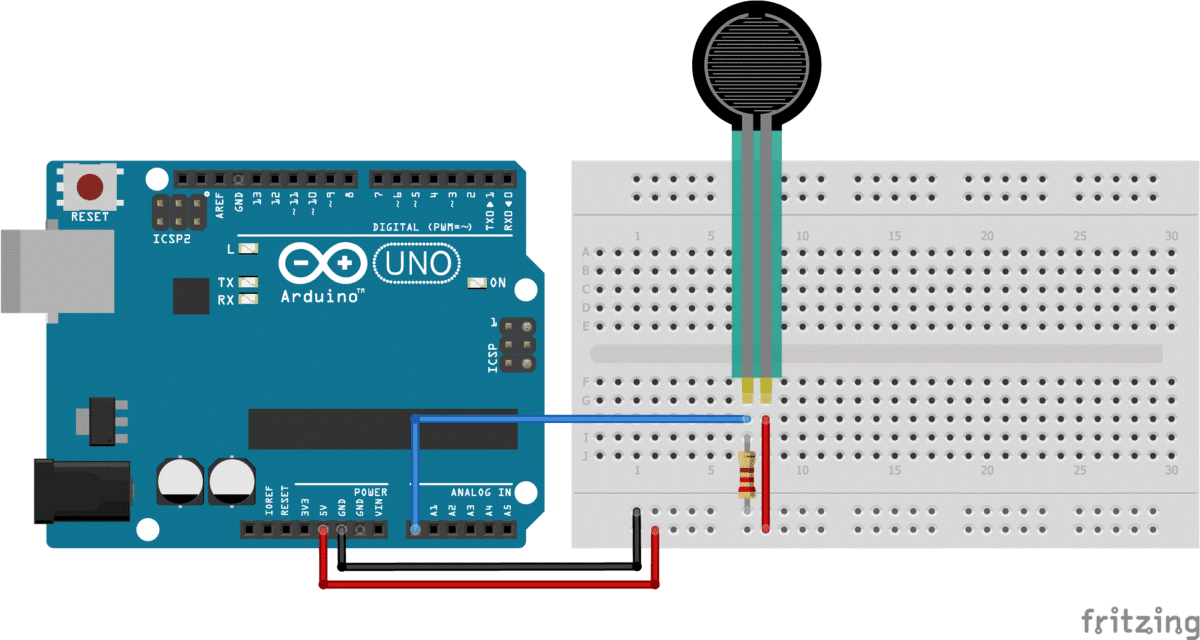}};
    \end{tikzpicture}
    \vspace{-5mm}
    \caption{(a) Force sensitive resistor (FSR): an off-the-shelf device (left) and the working principle of an FSR (right).
    (b) Circuit used for tactile sensing consisting of an FSR sensor, a resistor, and an Arduino UNO.
    }
    \label{fig:tactile_r}
    \vspace{-3mm}
\end{figure}

We used an off-the-shelf force sensitive sensor to detect the stiffness of the contacting material (Model Number: SEN-09375 ROHS). Figure \ref{fig:tactile_r} (a) shows the device on the left and the structure of the device on the right. The force sensor consists of three layers: a semi-conductive layer, a space, and a substrate with two interdigitating electrodes not overlapping. The device has an infinitely high output impedance if no force is applied. When a pressure is applied, the first layer and the third layer are partially contacted, causing the impedance of the device to change. 
% \vspace{-3mm}
% \begin{figure}[ht]
%     \centering
%     \includegraphics[trim={1.2cm 1.5cm 0 0},clip,width=0.485\textwidth]{fig/tactile_r.png}
%     \caption{
%     }
%     \label{fig:tactile_r}
% \end{figure}
The FSR device is connected with a reference resistor in series (Figure \ref{fig:tactile_r}~(b)). 
The output voltage from the circuit is
\vspace{-2mm}
$$ V_{out} = \frac{R_{ref}}{R_{ref}+R_{FSR}}V_{CC}$$
An Arduino UNO was used to control and gather data from the tactile sensor. 
% \begin{figure}[ht]
%     \centering
%     \includegraphics[trim={0 1.4cm 0 0},clip,width=0.485\textwidth]{fig/tactile_ardunio.png}
%     \caption{Circuit used for tactile sensing consisting of a FSR sensor, a resistor, and an Arduino UNO. 
%     }
%     \label{fig:tactile_circuit}
% \end{figure}
\vspace{-2mm}

Compliance is defined as the ratio of a deflection of the object over the applied force. In our implementation, we mount the tactile sensor on the rolling pin held in the robot gripper and program the robot to press against the contacting material and move a constant distance $\Delta x$ vertically while recording the reacting force $\Delta F$ generated during this process. 
For simplification, we always apply the force perpendicular to the contacting surface as the angle of the applied force also affects the detected compliance.
% detected under this setup.
% To simplify tasks, we always applied force perpendicular to the contacting surface. 

% From proposal but not sure it this is happening
% use the tactile information to ensure initial contact with the dough, continuing contact with the dough while rolling, and that sufficient force is exerted on the dough in order to distribute it from one place to another.

\vspace{-5mm}
\section{EXPERIMENTS} \label{sec:experiments}
We evaluated the roll start point methods (\textit{Centroid-2D}, \textit{Centroid-3D}, \textit{Highest-Point}), roll end point methods (\textit{Target}, \textit{Current}), and shrink action variants (\textit{Shrink-Disabled}, \textit{Forward-Shrink}, \textit{Side-Shrink}),
in the following experiments across three target shapes ($T_{3.5}$, $T_{4.0}$, $T_{4.5}$) and three materials (\textit{Play-Doh}, \textit{Plasticine}, \textit{Kinetic sand}). 
\setlength{\leftmargini}{20pt}
\begin{enumerate}[label=\Alph*)]
    \item Performance across materials (Section \ref{sec:experiments:materials} ~ )
    \item Roll start point methods (Section \ref{sec:experiments:start_method} ~ )
    \item Roll end point methods (Section \ref{sec:experiments:end_method} ~ )
    \item Shrink action (Section \ref{sec:experiments:shrink} ~ )
    \item Performance across target shapes (Section \ref{sec:experiments:target_shapes} ~ )
    \item Differentiable rendering (Section \ref{sec:experiments:differentiable_rendering} ~ )
    \item Tactile sensing (Section \ref{sec:experiments:tactile_sensing} ~ )
\end{enumerate}

We ran each experiment (A-E) $N = 3$ times 
and set the maximum time limit $T_{max} = 5~\mathrm{min}$ for each run. 
As evaluation metrics, we used: i) the intersection over union (IoU) between the current dough projected 2D shape and the target 2D shape, and ii) the maximum dough height obtained from the dough point cloud.

Video demonstrations of our experiments are available at \href{https://youtu.be/ZzLMxuITdt4}{https://youtu.be/ZzLMxuITdt4}

\subsection{Performance Across Materials}\label{sec:experiments:materials}
We investigated the differences between three dough-like deformable materials across  three roll start point methods. 
We set the target shape to the middle size $T_{4.0}$, roll end point method to \textit{Target}, and disabled the shrink action.

\begin{figure*}[ht]
    \centering
    \includegraphics[width=0.99\textwidth]{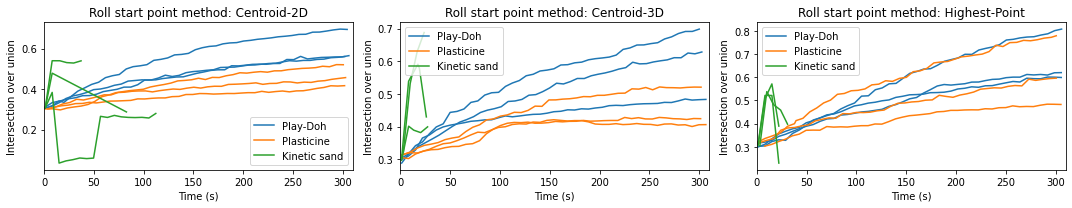}\vspace{-2mm}
    \caption{Intersection over union over time in terms of dough-like deformable
    materials (\textit{Play-Doh}, \textit{Plasticine}, \textit{Kinetic sand})
    for roll start point methods: \textit{Centroid-2D} (left), \textit{Centroid-3D} (middle), \textit{Highest-Point} (right). Three runs per condition.
    }
    \label{fig:experiments:materials_iou}
\end{figure*}

The results in terms of IoU over time are shown in Figure~\ref{fig:experiments:materials_iou}. 
We can observe that the runs with \textit{Play-Doh} tend to achieve the highest IoU for any start point method which suggests it is the easiest to expand to match the target shape. 
\textit{Plasticine} seems to be slightly more difficult to roll than \textit{Play-Doh}. 
The runs with \textit{Kinetic sand} were very unstable and we stopped the experiment if the material separated into several disconnected parts.

\begin{figure*}[ht]
    \centering
    \includegraphics[width=0.99\textwidth]{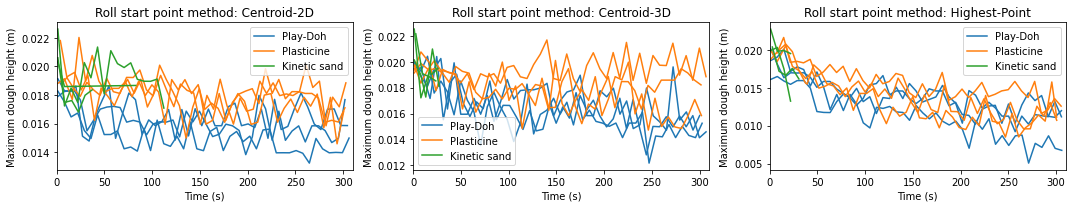}\vspace{-2mm}
    \caption{Maximum dough height in terms of dough-like deformable
    materials (\textit{Play-Doh}, \textit{Plasticine}, \textit{Kinetic sand})
    for each roll start point method: \textit{Centroid-2D} (left), \textit{Centroid-3D} (middle), \textit{Highest-Point} (right).
     Three runs per condition.
    % $N=3$ runs for each condition.
    }
    \label{fig:experiments:materials_height}
    \vspace{-3mm}
\end{figure*}

Figure~\ref{fig:experiments:materials_height} further shows the results in terms of the maximum dough height over time. 
We can see that the runs with \textit{Play-Doh} tend to achieve lower maximum dough heights which, similarly to the IoU metric, suggests it is the easiest to deform to match the target shape. 

Overall, these observations reflect the material properties described in Section \ref{sec:methods:hardware} ~. In further experiments we thus focused primarily on \textit{Play-Doh} as it was the best material to demonstrate the dough shaping.
\vspace{-1mm}

\subsection{Roll Start Point Methods}\label{sec:experiments:start_method}
We compared the performance of the three roll start point methods on \textit{Play-Doh} as well as \textit{Plasticine}. 
As shown in Section \ref{sec:experiments:materials} ~, the runs with \textit{Kinetic sand} were very short and unstable so we did not consider it further.
We set the target shape to the middle size $T_{4.0}$, roll end point method to \textit{Target}, and disabled the shrink action. 

\begin{figure}[ht]
    \centering
    \includegraphics[width=0.485\textwidth]{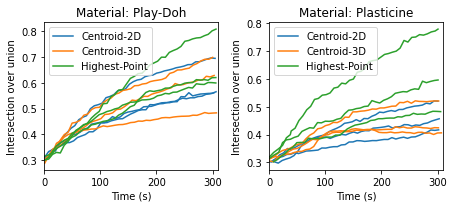}\vspace{-2mm}
    \caption{Intersection over union over time in terms~of roll start point method (\textit{Centroid-2D}, \textit{Centroid-3D}, \textit{Highest-Point}) for dough-like deformable materials \textit{Play-Doh} (left) and \textit{Plasticine} (right). 
    Three runs per condition.
    % $N=3$ runs for each condition.
    }%\vspace{-3mm}
    \label{fig:experiments:start_method}
    \vspace{-7mm}
\end{figure}

As can be seen from Figure~\ref{fig:experiments:start_method}, 
the \textit{Centroid-2D} method performed the worst in terms of IoU suggesting that the depth information that the \textit{Centroid-3D} and \textit{Highest-Point} methods utilize is important in determining the point from where to distribute the dough. 
The \textit{Highest-Point} method seems to be the best out of the three.
We hypothesize this can be explained by the observations that
the \textit{Centroid-3D} method was susceptible to converging to similar movements 
while the \textit{Highest-Point} method had higher variation in the start point location as the highest dough point was changing frequently. 
% Highest-point method performs the best
% Probably due to inherent variations in the start point location => it doesn’t get stuck doing similar movements
In the following experiments we thus used the \textit{Highest-Point} method.
\vspace{-1mm}

\subsection{Roll End Point Methods}\label{sec:experiments:end_method}
We further compared the performance of the \textit{Target} and \textit{Current} roll end point methods. We set the target shape to the largest size $T_{4.5}$ where the gap between the current and target dough shape is largest and so the differences between the \textit{Target} and \textit{Current} roll end point methods are most visible. We used the \textit{Play-Doh} material, the \textit{Highest-Point} roll start point method, and disabled the shrink action.

\begin{figure*}[ht]
    \centering
    \includegraphics[width=0.329\textwidth]{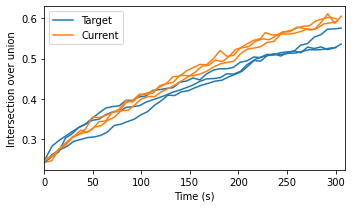}
    \includegraphics[width=0.329\textwidth]{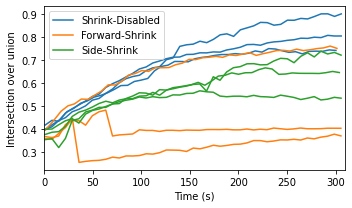}
    \includegraphics[width=0.329\textwidth]{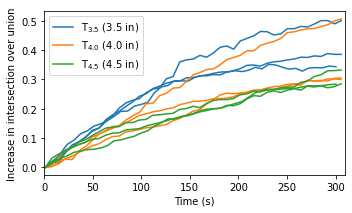}\vspace{-2mm}
    \caption{\hspace{-2mm}
    Intersection over union over time in terms of: roll end point method (left), shrink action setting (middle). Increase in intersection over union over time in terms of target shape size (circle diameters in inch) (right). 
    % $N=3$ runs for each condition.
     Three runs per condition.
    }
    \label{fig:experiments:end_method_and_shrink_and_targetshapes}
    \vspace{-3mm}
\end{figure*}

Our initial hypothesis that rolling beyond the dough boundary (determined at roll start) should expand the dough more towards the target shape as the dough boundary moves in that direction during the roll. Together with our preliminary experiments this suggested that the \textit{Target} method should achieve higher IoU faster which also informed the design of other experiments and the use of the \textit{Target} method.
However, as shown in Figure~\ref{fig:experiments:end_method_and_shrink_and_targetshapes}~(left), the \textit{Current} method achieves higher IoU, especially later during the run. 
One interpretation could be that the \textit{Current} method does more and shorter iterations in a given time, however, that was not the case as the total numbers of iterations of the Roll Dough Algorithm were very similar.
Therefore, we think a viable explanation could be that this is caused by the differences in the dynamics of the shorter vs.~longer roll movements. 
Indeed, humans would probably also stop the roll at (or close to) the dough boundary or at the instantaneous moving dough boundary, rather than rolling all the way to the target shape and rolling on a table where there is no dough.
It would be thus interesting to further experiment with a method that ends the roll at the instantaneous moving dough boundary which would require a camera mounted on the end-effector.

\subsection{Shrink Action}\label{sec:experiments:shrink}
% \vspace{-1mm}
We investigated the influence of enabling the shrink action and evaluated the performance in terms of IoU in three settings: \textit{Shrink-Disabled}, \textit{Forward-Shrink}, and \textit{Side-Shrink}. To make valid comparisons, when the shrink action was enabled we considered only the runs where the shrink action was eventually used. 
We also set a threshold (in terms of the width of the target outline) to prevent triggering the shrink action in case of small dough expansions beyond the target shape outline.
We set the target shape to the smallest size $T_{3.5}$ when it is easiest to expand or move the dough outside of the target shape and so the differences between the shrink settings are most visible. We used the \textit{Play-Doh} material, the \textit{Highest-Point} roll start point method, and the \textit{Target} end point method.

Figure~\ref{fig:experiments:end_method_and_shrink_and_targetshapes}~(middle) shows that the \textit{Shrink-Disabled} setting achieves higher IoU than either of the settings with shrink action enabled. 
As can be seen from the four worst runs with shrink action enabled, once the shrink action was performed the IoU plateaued afterwards. 
This suggests that there is a tradeoff between dough expansion and shrinking and needs to be tuned.
For instance, the shrink action should be allowed only if the estimated gains in IoU from its execution are higher than those from the expand action. 
Also, it might be better to enable the shrink action towards the end of the run, for example, once a certain minimum IoU is achieved.

Comparing \textit{Forward-Shrink} and \textit{Side-Shrink}, the \textit{Forward-Shrink} setting performed much worse once the shrink action was executed. 
Indeed, pushing the dough with the rolling side is not very efficient as it likely results in a roll over the dough. The \textit{Side-Shrink} provides a much more rigid push. 

% \begin{figure}[ht]
%     \centering
%     \includegraphics[width=0.485\textwidth]{fig/experiments/shrink_iou_playdoh.png}
%     \caption{
%     Intersection over union over time in terms of shrink action setting: \textit{Shrink-Disabled}, \textit{Forward-Shrink}, and \textit{Side-Shrink}. 
%     $N=3$ runs for each condition.
%     }
%     \label{fig:experiments:shrink}
% \end{figure}

\subsection{Performance Across Target Shapes}\label{sec:experiments:target_shapes}
We also examined the impact of the target shape size on performance. 
To make valid comparisons between different target shape sizes, we evaluated the increase in IoU with respect to the initial IoU. 
We used the \textit{Play-Doh} material, the \textit{Highest-Point} roll start point method, the \textit{Target} end point method, and disabled the shrink action.

As shown in Figure~\ref{fig:experiments:end_method_and_shrink_and_targetshapes}~(right), for smaller target shapes the increase in IoU can be achieved faster. 
Since we used the \textit{Target} end point method
this might just reflect the observations from Section \ref{sec:experiments:end_method} ~ 
as rolling only to the current shape boundary is more similar to rolling to the target shape outline for smaller target shapes than for larger target shapes. 
It would be thus interesting to further repeat this experiment with the \textit{Current} end point method.
In this experiment, we also recorded the highest IOU exceeding 0.90.

% \begin{figure}[ht]
%     \centering
%     \includegraphics[width=0.485\textwidth]{fig/experiments/targetshapes_iou_playdoh.png}
%     \caption{
%     Increase in intersection over union over time in terms of target shape size: $T_{3.5}$, $T_{4.0}$, and $T_{4.5}$ (circle diameters in inch). 
%     $N=3$ runs for each condition.
%     }
%     \label{fig:experiments:target_shape}
% \end{figure}

\subsection{Differentiable Rendering}\label{sec:experiments:differentiable_rendering}
\begin{figure}[ht]
    \centering
    \includegraphics[width=0.15\textwidth]{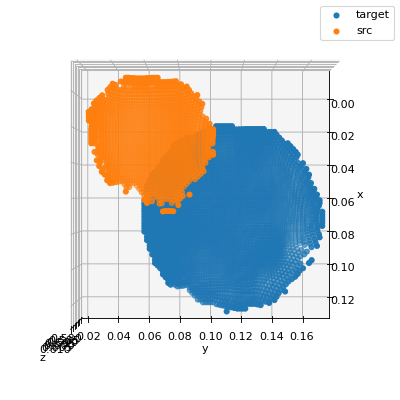}
    \includegraphics[width=0.15\textwidth]{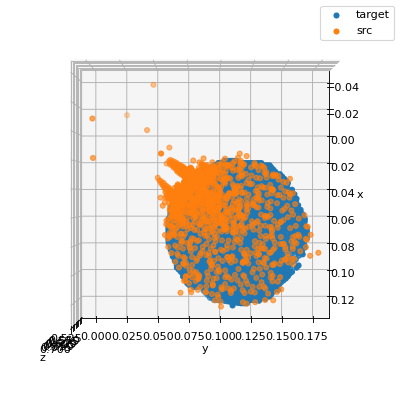}
    \includegraphics[width=0.15\textwidth]{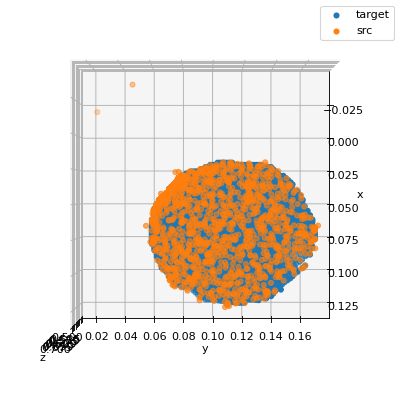}
    \includegraphics[width=0.45\textwidth]{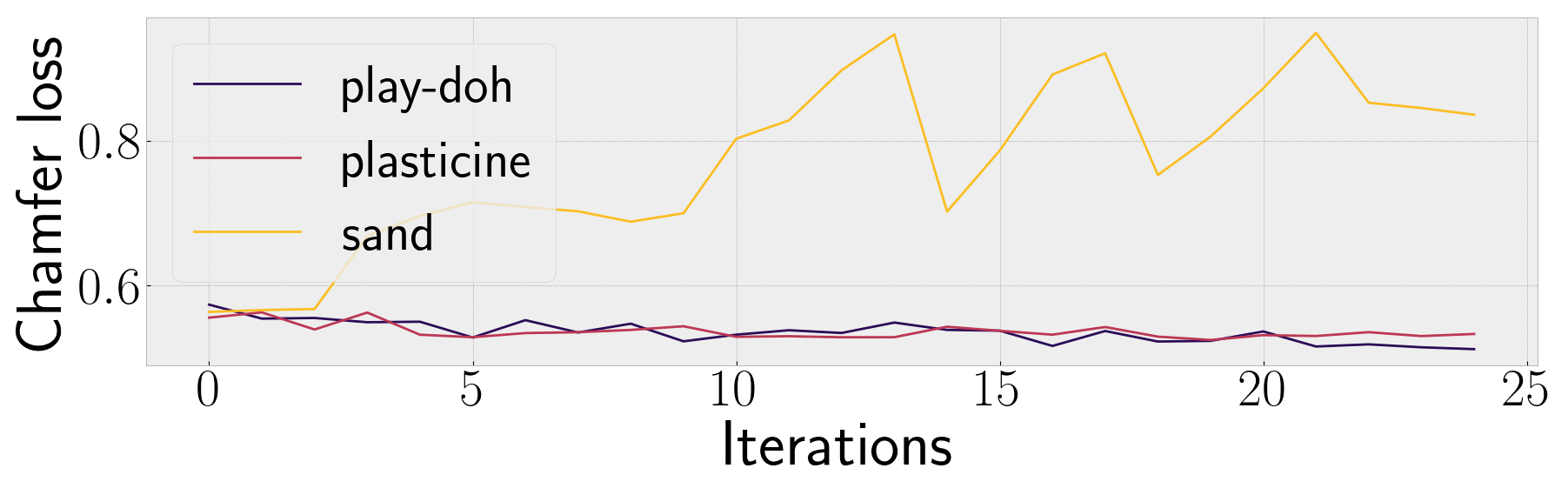}
    \vspace{-1mm}
    \caption{\textbf{Top:} Deforming the source shape to target shape using SGD optimizer with differentiable point-cloud rendering. 
    \textbf{Bottom:} Real robot experiment with \textit{Play-Doh}, \textit{Plasticine}, \textit{Kinetic sand}, showing the Chamfer distance between current and target dough shape at each iteration. }
    \vspace{-5mm}
    \label{fig:experiments:diff_syn}
\end{figure}

We tested the differentiable method by simulating synthetic deformation of the dough point cloud to the target shape, and ran real robot experiments towards the target shape $T_{4.5}$ for each material. The change of Chamfer loss after across iterations is shown in \autoref{fig:experiments:diff_syn}.
% Here we demonstrate 
With these preliminary results we demonstrate the use of differentiable rendering for dough shaping task and further experiments are needed to fully evaluate this approach.

\subsection{Tactile Sensing}\label{sec:experiments:tactile_sensing}
\begin{figure}[ht]
    \centering
    \hspace*{-1.1ex}%
    \begin{tikzpicture}[x=0.5\textwidth,y=0.1\textwidth,every text node part/.style={align=center}]
    \node at (-0.1, 0.0){ };
    \node at (0.0, -0.89){(a)};
    \node at (0.5, -0.89){(b)};
    \node at (0.46, 0.0){\includegraphics[width=0.24\textwidth]{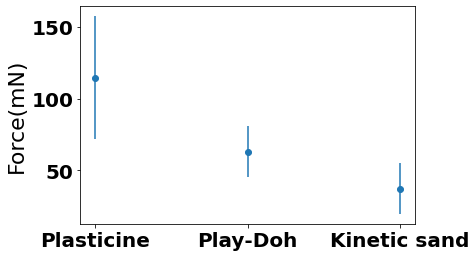}};
    \node at (0.0, 0.0){\includegraphics[width=0.215\textwidth]{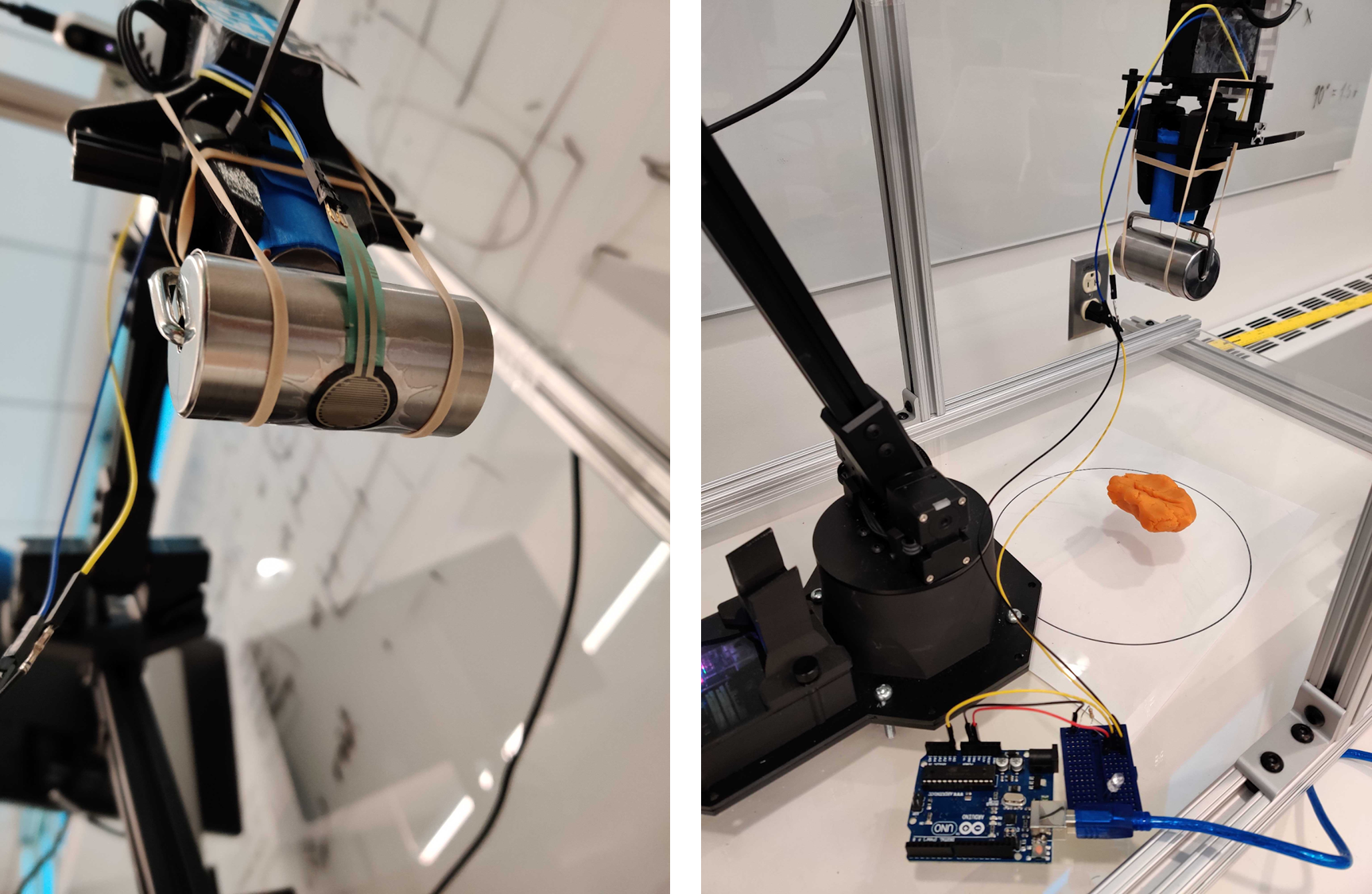}};
    \end{tikzpicture}
    \vspace{-2mm}
    \caption{(a) Experimental setup for tactile sensing: tactile sensor glued on the rolling pin (left), robot gripper controlled to apply force perpendicular to the dough surface (right).
    (b) Resisting forces of the three deformable materials.
    }
    \label{fig:tactile_exp}
    \vspace{-6mm}
\end{figure}
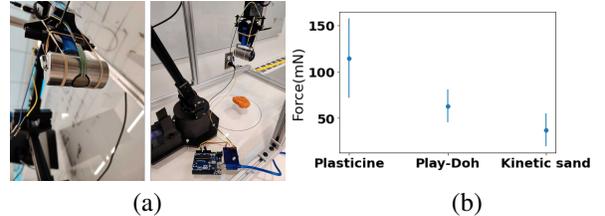

We used tactile sensing to detect the stiffness of the three materials introduced above. The information obtained here can be used in future to determine the force applied when rolling different materials. 
Figure~\ref{fig:tactile_exp}~(a) shows the experimental setup for tactile sensing. 
% We programmed the robot gripper to move down a constant distance and recorded 5 measurements for each material. 
We programmed the robot to push the dough a fixed vertical distance and recorded 5 measurements for each material. 
Before each measurement we reshaped the dough into the initial shape using a mold.
% Figure~\ref{fig:tactile_exp}~(b) shows the resisting force measured by the sensor. 
% The smaller the resisting force is, the softer the material is. 
% \textit{Play-Doh} turned out to have an appropriate stiffness for rolling experiments. 
As shown in Figure~\ref{fig:tactile_exp}~(b), the resisting forces differ considerably across materials (smaller force implies softer material).
This suggests the material stiffness measurements can help the dough shaping control policy adapt to different materials.
% This suggests the use of the material stiffness measurements prior and/or during shaping 
% The relatively large variance shown in Figure~\ref{fig:tactile_exp}~(b) was likely caused by the uneven surfaces of the dough-like materials.
% \vspace{-2mm}
\vspace{-4mm}
\section{CONCLUSION} \label{sec:conclusion}
In this paper, we addressed the problem of shaping a piece of dough-like deformable material into a predefined 2D target shape. 
We used a 6 degree-of-freedom WidowX-250 Robot Arm equipped with a rolling pin and information collected from an RGB-D camera and a tactile sensor. 

We developed a \href{https://github.com/jancio/Robotic-Dough-Shaping}{Roll Dough GUI Application} and proposed several control policies, including a dough shrinking action. 
We evaluated these policies in extensive experiments across three kinds of deformable materials and across three target dough shapes, exceeding the intersection over union (IoU) of 0.90. 
% Our Roll Dough GUI Application

Our results show that: i) rolling dough from the highest dough point is more efficient than from the 2D/3D dough centroid; ii) it might be better to stop the roll movement at the current dough boundary as opposed to the target shape outline; iii) the shrink action might be beneficial only if properly tuned with respect to the expand action; and iv) the \textit{Play-Doh} material is easier to shape to a target shape as compared to \textit{Plasticine} or \textit{Kinetic sand}.

Video demonstrations of our work are available at \href{https://youtu.be/ZzLMxuITdt4}{https://youtu.be/ZzLMxuITdt4}

\vspace{-4mm}
\section{FUTURE WORK} \label{sec:future_work}

% In future work, we plan to explore the following research directions.
The following ideas can be investigated as future work.

\textbf{Evaluation Metrics.}
Instead of the IoU metric evaluating the projection of the current dough shape, employ a 3D evaluation metric, for example, to explicitly optimize for even dough height distribution.

\textbf{Target Shapes.}
Evaluate generalizability of our method to different kinds of target shapes such as ellipses.

\textbf{Materials.}
Choose a set of dough-like deformable materials with intentionally 
different quantitative physical properties to demonstrate generalizability of our method to different materials.
Also, as we observed, \textit{Play-Doh} hardens over time and becomes difficult to spread/roll. It might be thus useful to account for such changes in material properties over time and develop algorithms that adjust to time-evolving materials.

\textbf{Policy Learning and Tactile Sensing.} 
% Instead of using criteria or differentiable physics simulation for motion planning, predefine some basics movement and train networks for robot to operated the object with some combination of the basic movements. 
Learn the dough rolling policy online and/or from demonstrations. 
Incorporate stiffness measurements from dough tactile sensing as a prior for learning. 
Also, the tactile sensor can be attached just above the rolling pin so that it can get in contact with the pin but does not prevent rolling. The stiffness measurements can be then used during the roll action, for example, to adjust the rolling pin height, and to determine the moment the dough was touched.

\textbf{Roll Start and End Point Methods.}
Experiment with adding little random noise to the selection of the roll start point, based on observations in Section \ref{sec:experiments:start_method} ~. 
% \paragraph{Roll End Point Method}
Experiment with a new roll end point method that ends the roll within a certain tolerance of the instantaneous moving dough boundary which would require a camera mounted on the end-effector, as suggested in Section \ref{sec:experiments:end_method} ~.

% Further suggestions for future work are provided in Section \ref{sec:appendix:future_work} ~.
See \ref{sec:appendix:future_work}~ ~ for further future work suggestions.
\vspace{-7mm}
% \input{sections/acknowledgement}

%\section*{ACKNOWLEDGEMENT}
%
%This paper has been supported by NRF of Korea in 2021.

%%%%%%%%%%%%%%%%% BIBLIOGRAPHY IN THE LaTeX file !!!!! %%%%%%%%%%%%%%%%%%%%%%
%%---------------------------------------------------------------------------%%
\bibliographystyle{unsrt}
\bibliography{references}

\begin{thebibliography}{10}

\bibitem{hui2017visual}
Fei Hui, Pierre Payeur, and Ana-Maria Cretu.
\newblock Visual tracking of deformation and classification of non-rigid
  objects with robot hand probing.
\newblock {\em Robotics}, 6(1):5, 2017.

\bibitem{guler2015estimating}
Puren Guler, Karl Pauwels, Alessandro Pieropan, Hedvig Kjellstr{\"o}m, and
  Danica Kragic.
\newblock Estimating the deformability of elastic materials using optical flow
  and position-based dynamics.
\newblock In {\em 2015 IEEE-RAS 15th International Conference on Humanoid
  Robots (Humanoids)}, pages 965--971. IEEE, 2015.

\bibitem{nair2017combining}
Ashvin Nair, Dian Chen, Pulkit Agrawal, Phillip Isola, Pieter Abbeel, Jitendra
  Malik, and Sergey Levine.
\newblock Combining self-supervised learning and imitation for vision-based
  rope manipulation.
\newblock In {\em 2017 IEEE international conference on robotics and automation
  (ICRA)}, pages 2146--2153. IEEE, 2017.

\bibitem{yan2020learning}
Wilson Yan, Ashwin Vangipuram, Pieter Abbeel, and Lerrel Pinto.
\newblock Learning predictive representations for deformable objects using
  contrastive estimation.
\newblock {\em arXiv preprint arXiv:2003.05436}, 2020.

\bibitem{rouhafzay2020transfer}
Ghazal Rouhafzay, Ana-Maria Cretu, and Pierre Payeur.
\newblock Transfer of learning from vision to touch: a hybrid deep
  convolutional neural network for visuo-tactile 3d object recognition.
\newblock {\em Sensors}, 21(1):113, 2020.

\bibitem{li2015regrasping}
Yinxiao Li, Danfei Xu, Yonghao Yue, Yan Wang, Shih-Fu Chang, Eitan Grinspun,
  and Peter~K Allen.
\newblock Regrasping and unfolding of garments using predictive thin shell
  modeling.
\newblock In {\em 2015 IEEE International Conference on Robotics and Automation
  (ICRA)}, pages 1382--1388. IEEE, 2015.

\bibitem{zhu2021vision}
Jihong Zhu, David Navarro-Alarcon, Robin Passama, and Andrea Cherubini.
\newblock Vision-based manipulation of deformable and rigid objects using
  subspace projections of 2d contours.
\newblock {\em Robotics and Autonomous Systems}, 142:103798, 2021.

\bibitem{navarro2017fourier}
David Navarro-Alarcon and Yun-Hui Liu.
\newblock Fourier-based shape servoing: A new feedback method to actively
  deform soft objects into desired 2-d image contours.
\newblock {\em IEEE Transactions on Robotics}, 34(1):272--279, 2017.

\bibitem{lagneau2020automatic}
Romain Lagneau, Alexandre Krupa, and Maud Marchal.
\newblock Automatic shape control of deformable wires based on model-free
  visual servoing.
\newblock {\em IEEE Robotics and Automation Letters}, 5(4):5252--5259, 2020.

\bibitem{mcconachie2020manipulating}
Dale McConachie, Andrew Dobson, Mengyao Ruan, and Dmitry Berenson.
\newblock Manipulating deformable objects by interleaving prediction, planning,
  and control.
\newblock {\em The International Journal of Robotics Research}, 39(8):957--982,
  2020.

\bibitem{ramirez2014motion}
Ixchel~G Ramirez-Alpizar, Kensuke Harada, and Eiichi Yoshida.
\newblock Motion planning for dual-arm assembly of ring-shaped elastic objects.
\newblock In {\em 2014 IEEE-RAS International Conference on Humanoid Robots},
  pages 594--600. IEEE, 2014.

\bibitem{alonso2015local}
Javier Alonso-Mora, Ross Knepper, Roland Siegwart, and Daniela Rus.
\newblock Local motion planning for collaborative multi-robot manipulation of
  deformable objects.
\newblock In {\em 2015 IEEE international conference on robotics and automation
  (ICRA)}, pages 5495--5502. IEEE, 2015.

\bibitem{mitsioni2019data}
Ioanna Mitsioni, Yiannis Karayiannidis, Johannes~A Stork, and Danica Kragic.
\newblock Data-driven model predictive control for the contact-rich task of
  food cutting.
\newblock In {\em 2019 IEEE-RAS 19th International Conference on Humanoid
  Robots (Humanoids)}, pages 244--250. IEEE, 2019.

\bibitem{su2019improved}
Hang Su, Chenguang Yang, Giancarlo Ferrigno, and Elena De~Momi.
\newblock Improved human--robot collaborative control of redundant robot for
  teleoperated minimally invasive surgery.
\newblock {\em IEEE Robotics and Automation Letters}, 4(2):1447--1453, 2019.

\bibitem{sundaresan2020learning}
Priya Sundaresan, Jennifer Grannen, Brijen Thananjeyan, Ashwin Balakrishna,
  Michael Laskey, Kevin Stone, Joseph~E Gonzalez, and Ken Goldberg.
\newblock Learning rope manipulation policies using dense object descriptors
  trained on synthetic depth data.
\newblock In {\em 2020 IEEE International Conference on Robotics and Automation
  (ICRA)}, pages 9411--9418. IEEE, 2020.

\bibitem{zhu2018dual}
Jihong Zhu, Benjamin Navarro, Philippe Fraisse, Andr{\'e} Crosnier, and Andrea
  Cherubini.
\newblock Dual-arm robotic manipulation of flexible cables.
\newblock In {\em 2018 IEEE/RSJ International Conference on Intelligent Robots
  and Systems (IROS)}, pages 479--484. IEEE, 2018.

\bibitem{chi2022iterative}
Cheng Chi, Benjamin Burchfiel, Eric Cousineau, Siyuan Feng, and Shuran Song.
\newblock Iterative residual policy: for goal-conditioned dynamic manipulation
  of deformable objects.
\newblock {\em arXiv preprint arXiv:2203.00663}, 2022.

\bibitem{ha2022flingbot}
Huy Ha and Shuran Song.
\newblock Flingbot: The unreasonable effectiveness of dynamic manipulation for
  cloth unfolding.
\newblock In {\em Conference on Robot Learning}, pages 24--33. PMLR, 2022.

\bibitem{thananjeyan2017multilateral}
Brijen Thananjeyan, Animesh Garg, Sanjay Krishnan, Carolyn Chen, Lauren Miller,
  and Ken Goldberg.
\newblock Multilateral surgical pattern cutting in 2d orthotropic gauze with
  deep reinforcement learning policies for tensioning.
\newblock In {\em 2017 IEEE International Conference on Robotics and Automation
  (ICRA)}, pages 2371--2378. IEEE, 2017.

\bibitem{huang2021plasticinelab}
Zhiao Huang, Yuanming Hu, Tao Du, Siyuan Zhou, Hao Su, Joshua~B Tenenbaum, and
  Chuang Gan.
\newblock Plasticinelab: A soft-body manipulation benchmark with differentiable
  physics.
\newblock {\em arXiv preprint arXiv:2104.03311}, 2021.

\bibitem{matl2021deformable}
Carolyn Matl and Ruzena Bajcsy.
\newblock Deformable elasto-plastic object shaping using an elastic hand and
  model-based reinforcement learning.
\newblock In {\em 2021 IEEE/RSJ International Conference on Intelligent Robots
  and Systems (IROS)}, pages 3955--3962. IEEE, 2021.

\bibitem{nicholls1989survey}
Howard~R Nicholls and Mark~H Lee.
\newblock A survey of robot tactile sensing technology.
\newblock {\em The International Journal of Robotics Research}, 8(3):3--30,
  1989.

\bibitem{xu2013tactile}
Danfei Xu, Gerald~E Loeb, and Jeremy~A Fishel.
\newblock Tactile identification of objects using bayesian exploration.
\newblock In {\em 2013 IEEE International Conference on Robotics and
  Automation}, pages 3056--3061. IEEE, 2013.

\bibitem{allen1988integrating}
Peter~K Allen.
\newblock Integrating vision and touch for object recognition tasks.
\newblock {\em The International Journal of Robotics Research}, 7(6):15--33,
  1988.

\bibitem{allen1988haptic}
P.K. Allen and K.S. Roberts.
\newblock Haptic object recognition using a multi-fingered dextrous hand.
\newblock In {\em Proceedings, 1989 International Conference on Robotics and
  Automation}, pages 342--347 vol.1, 1989.

\bibitem{jamali2011majority}
Nawid Jamali and Claude Sammut.
\newblock Majority voting: Material classification by tactile sensing using
  surface texture.
\newblock {\em IEEE Transactions on Robotics}, 27(3):508--521, 2011.

\bibitem{fazeli2019see}
Nima Fazeli, Miquel Oller, Jiajun Wu, Zheng Wu, Joshua~B Tenenbaum, and Alberto
  Rodriguez.
\newblock See, feel, act: Hierarchical learning for complex manipulation skills
  with multisensory fusion.
\newblock {\em Science Robotics}, 4(26):eaav3123, 2019.

\bibitem{romano2011human}
Joseph~M Romano, Kaijen Hsiao, G{\"u}nter Niemeyer, Sachin Chitta, and
  Katherine~J Kuchenbecker.
\newblock Human-inspired robotic grasp control with tactile sensing.
\newblock {\em IEEE Transactions on Robotics}, 27(6):1067--1079, 2011.

\bibitem{lee2020making}
Michelle~A Lee, Yuke Zhu, Peter Zachares, Matthew Tan, Krishnan Srinivasan,
  Silvio Savarese, Li~Fei-Fei, Animesh Garg, and Jeannette Bohg.
\newblock Making sense of vision and touch: Learning multimodal representations
  for contact-rich tasks.
\newblock {\em IEEE Transactions on Robotics}, 36(3):582--596, 2020.

\bibitem{opencvlibrary}
G.~Bradski.
\newblock {The OpenCV Library}.
\newblock {\em Dr. Dobb's Journal of Software Tools}, 2000.

\bibitem{wu2021densityaware}
Tong Wu, Liang Pan, Junzhe Zhang, Tai WANG, Ziwei Liu, and Dahua Lin.
\newblock Density-aware chamfer distance as a comprehensive metric for point
  cloud completion.
\newblock In {\em In Advances in Neural Information Processing Systems
  (NeurIPS), 2021}, 2021.

\bibitem{Yifan:DSS:2019}
Wang Yifan, Felice Serena, Shihao Wu, Cengiz {\"{O}}ztireli, and Olga
  Sorkine{-}Hornung.
\newblock Differentiable surface splatting for point-based geometry processing.
\newblock {\em ACM Transactions on Graphics (proceedings of ACM SIGGRAPH
  ASIA)}, 38(6), 2019.

\bibitem{hu2019difftaichi}
Yuanming Hu, Luke Anderson, Tzu-Mao Li, Qi~Sun, Nathan Carr, Jonathan
  Ragan-Kelley, and Fr{\'e}do Durand.
\newblock Difftaichi: Differentiable programming for physical simulation.
\newblock {\em ICLR}, 2020.

\end{thebibliography}
% \clearpage
% \newpage
\vspace{-5mm}
\section{APPENDIX}\label{sec:appendix}
\vspace{-1mm}
% \section*{APPENDIX}\label{sec:appendix}
% possibly without section number
\subsection{Roll Dough GUI Application}\label{sec:appendix:app}
\vspace{-2mm}
\begin{figure}[h!]
    \centering
    \includegraphics[width=0.465\textwidth]{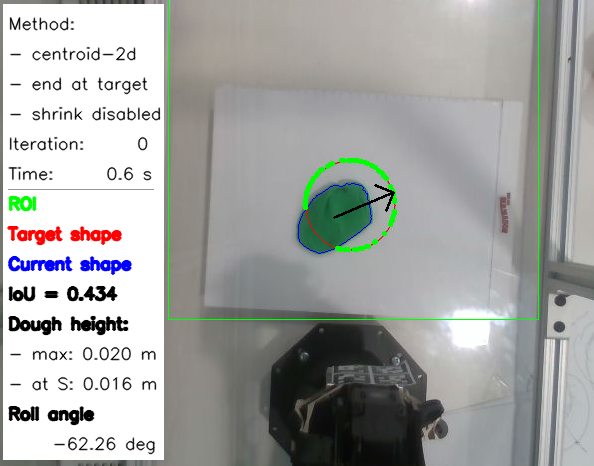}
    \caption{Screenshot of Roll Dough GUI Application available at 
\href{https://github.com/jancio/Robotic-Dough-Shaping}{https://github.com/jancio/Robotic-Dough-Shaping}}
    \label{fig:appendix:app}
    \vspace{-3mm}
\end{figure}

\subsection{Further Future Work Suggestions}\label{sec:appendix:future_work}

\textbf{Shrink Action.}
Examine the tradeoff between the expand and shrink action, as suggested in Section \ref{sec:experiments:shrink} ~, for instance, the shrink action can be allowed only if the estimated gains in IoU from its execution exceed those from the expand action. 
Also, evaluate a more realistic scenario when two tools are used: one for the expand action and the other for the shrink action.

\textbf{Physical Model.}
Currently in our work, we are not using any physical model of the plastic deformation. One exciting future direction is to apply a physical model and differentiate it to get the derivatives to aid the planning. However, since we have access to the dough point cloud, we can apply physically based differentiable "rendering" to get the gradient which aids in planing of the roll. Our current work does not consider the appearance of the material which could incorporate the surface normal information and the material reflectance (for example, color information). It is also interesting to apply differentiable surface splatting~\cite{Yifan:DSS:2019} to compute the gradient. Besides the differentiable visual perceptions, we could also differentiate the physical process with simulator \cite{hu2019difftaichi}, which can better predict object deformation after applying force.

\textbf{Robotic Platforms.}
Demonstrate generalizability of our method to various robotic platforms, for example, with the Kinova Gen3 robotic arm.

\end{document}